\documentclass[conference]{IEEEtran}
\IEEEoverridecommandlockouts
\usepackage{cite}
\usepackage{amsmath,amssymb,amsfonts}
\usepackage{algorithmic}
\usepackage{graphicx}
\usepackage{textcomp}
\usepackage{xcolor}

\newcommand\T{\rule{0pt}{2.6ex}}       

\def\BibTeX{{\rm B\kern-.05em{\sc i\kern-.025em b}\kern-.08em
    T\kern-.1667em\lower.7ex\hbox{E}\kern-.125emX}}
\begin{document}
\title{Discriminating Original Region from Duplicated One in Copy-Move Forgery\\
}
\author{\IEEEauthorblockN{ Saba Salehi}
\IEEEauthorblockA{\textit{Cyberspace Research Inst.} \\
\textit{Shahid Beheshti University}\\
Tehran, Iran \\
sab.salehi@mail.sbu.ac.ir}
\and
\IEEEauthorblockN{ Ahmad Mahmoodi-Aznaveh}
\IEEEauthorblockA{\textit{Cyberspace Research Inst.} \\
\textit{Shahid Beheshti University}\\
Tehran, Iran \\
a\_mahmoudi@sbu.ac.ir}
}
\maketitle
\begin{abstract}
Since images are used as evidence in many cases, validation of digital images is essential. Copy-move forgery is a special kind of manipulation in which some parts of an image is copied and pasted into another part of the same image. Various methods have been proposed to detect copy-move forgery, which have achieved promising results. In previous methods, a binary mask determining the original and forged region is presented as the final result. However, it is not specified which part of the mask is the forged region. It should be noted that discriminating the original region from the duplicated one is not usually feasible by human visual system(HVS). On the other hand, exact localizing the forged region can be helpful for automatic forgery detection especially in combined forgeries. In real-world forgeries, some manipulations are performed in order to provide a visibly realistic scene. These modifications are usually applied on the boundary of the duplicated snippets. In this research, the texture information of the border regions of both the original and copied patches have been statistically investigated. Based on this analysis, we propose a method to discriminated copied snippets from original ones. In order to validate our method, GRIP dataset is utilized since it contains more realistic forged images which are not easily recognizable by HVS.

\end{abstract}


\section{Introduction}\label{Intro}
With the development of advanced image editing software, image forging can be performed easily. Since the images are usually used as evidence, the authentication of digital images is important. Manipulation detection methods are divided into active and passive approaches. The active approach, including digital signature and digital watermarking, requires extra information to detect manipulations. Therefore, they could not be used in all applications. Consequently, the passive approach was proposed, which mainly seeks to find out the statistical inconsistency in natural images.

\begin{figure}[htbp]
\centerline{\includegraphics[width=0.5\textwidth]{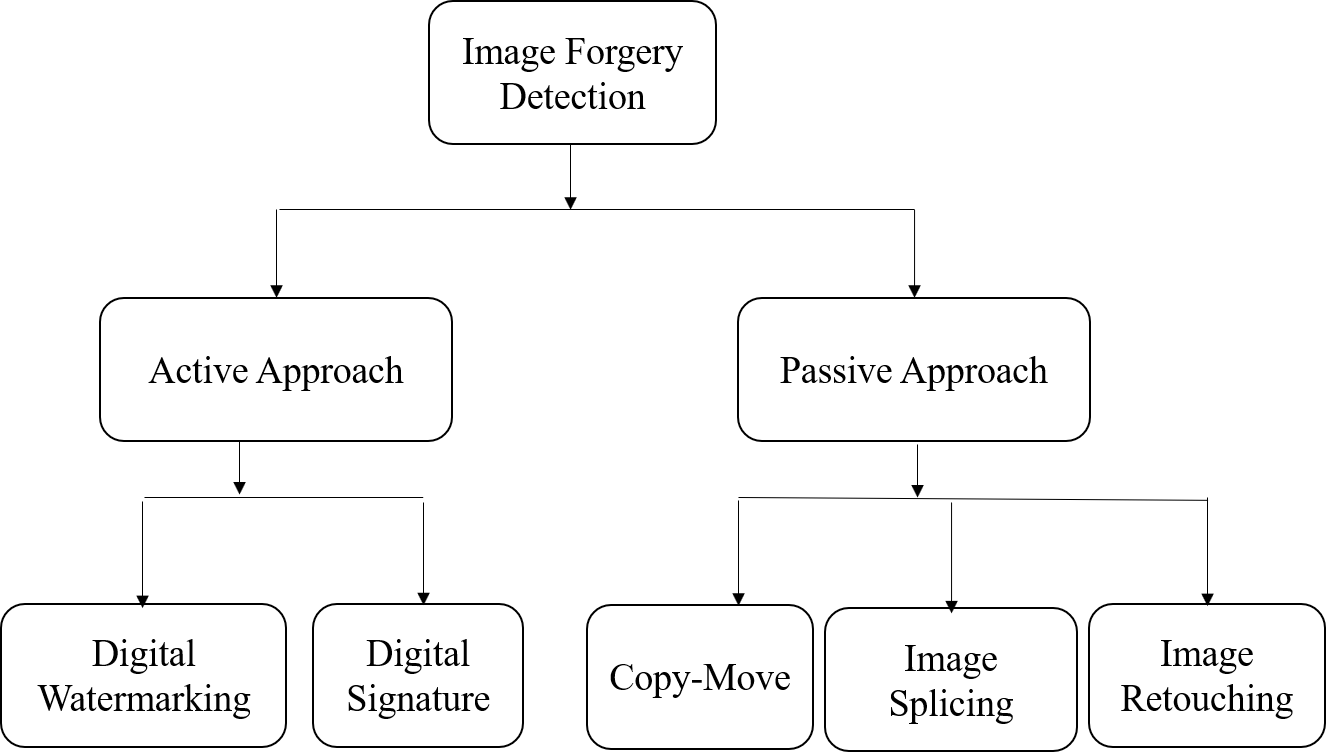}}
\caption{Classification of Image Forgery Detection Techniques.}
\label{class.forg}
\end{figure}

Image manipulation can be generally classified into three types\cite{boididou2018verifying}, image retouching, image splicing, copy-move forgery. Image retouching, improving image appearance such as contrast, is the least harmful kind of manipulation. A forged image is created by combining more than two images in splicing. In copy-move forgery, some snippets of an image are copied and pasted to other parts of the same image. This forgery is usually used to hide an object or increase the number of objects for exaggerating in the image (fig.\ref{class.forg}).

Splicing may introduce inconsistencies in image characteristics; hence, splicing detection methods are mostly based on analyzing the inconsistencies among local features. Since in copy-move forgery copied region is part of the original image, it is not feasible to employ splicing detection method to such images. Copy-move forgery detection methods are based on finding similarity in an image.

Numerous methods have been proposed to detect copy-move forgery. Some of them present impressive results. However, none of them discriminate the original and forged region. They only provide a binary mask containing the original and duplicated regions. In this paper, we present a method which can discriminate original patches from forged ones for the first time. It should be noted that it is a challenging task due to the similarity of image properties in the original and duplicated regions.

This paper is organized as follows: In section \ref{review}, a review of forgery detection methods is presented. We will examine the proposed method in section \ref{method}. The proposed method is evaluated in section \ref{result}. Finally, we clarify our conclusion.

\section{Review of Passive Detection Approach in Forgery Images}\label{review}

\begin{figure*}[htbp]
\centerline{\includegraphics[width=1\textwidth]{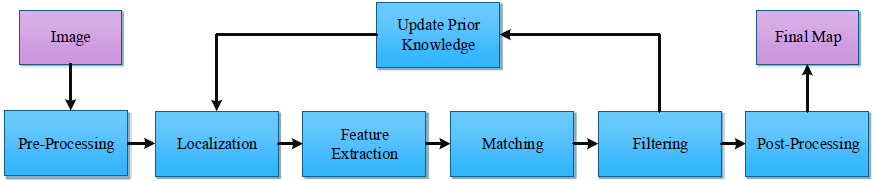}}
\caption{Common framework of the copy-move forgery detection methods\cite{zandi2016iterative}.}
\label{CMFD}
\end{figure*}

The most common way of manipulating images is splicing forgery. It is combines two (or more) images to create tampered images. Forgery detection methods are mostly based on inconsistency in image characteristics. As a case in point, photos captured by different digital cameras usually have different noise pattern. In this case, forged images which created with the different camera can be detected by investigating the noise \cite{pun2016multi}. Another technique for splicing forgery detection is performed by analyzing the amount and types of blurriness such as motion and out-of-focus blur \cite{bahrami2014image}.

In copy-move forgery, one or more image regions are copied and pasted in another region of the same image. Since the copied one is from the original image itself, some image properties, such as noise, are identical in the original and copied region. Therefore, methods used to detect splicing forgery are not usable in copy-move forgery. The main approach for detecting copy-move forgery is to identify similar regions. To detect this kind of forgery, several methods have been proposed, which follow a general framework that is shown in Fig.\ref{CMFD}.

There are some preprocessing phases in the first step to facilitate the next steps. Converting color image to grayscale image is one of them. The second step is localization. This step can be implemented with two approaches: block-based and keypoint-based. The block-based approach divides the image into overlapping blocks, and all possible blocks are extracted. In keypoint-based approach, high entropy regions i.e. keypoints are just explored. In the feature extraction step, using a descriptor, features are extracted from each block or keypoint. Afterward, a feature vector is obtained. Feature vectors are compared to find similar blocks. Due to self-similarity property of natural images, a number of similar blocks are detected incorrectly.
In the next step, the matched pairs are filtered to remove the blocks that are matched incorrectly.
 In the step of updating information, some methods repeat some previous steps in order to more precisely localize forged regions. Processing steps are also are used to make the output more accurate. The resultant output in copy-move forgery detection methods are a binary mask in which the forged region is not discriminated from the original one.

The first method by Friedrich et al. \cite{fridrich2003detection} proposed to detect copy-move forgery. First, they examine all of the overlapping image blocks. Then, discrete cosine transform (DCT) is used for describing overlapped blocks. The features extracted from each block are lexicographically arranged. Then, the similar blocks are found by comparing their description. Since truly matched pairs have a same shift vector, it is used to remove the falsely matched pairs. Other block-based methods work generally like the mentioned method, with some improvements For example, in feature extraction, they use rotation invariant transforms. \cite{li2013image} \cite{ryu2013rotation}, Some methods use hashing -based matching algorithms such as LSH \cite{li2013image},\cite{ryu2013rotation}. Another approach is to use improved filtering algorithms \cite{li2013image} \cite{ryu2013rotation}. 

Block-based methods have heavy computations because all image blocks have considered. Hence, keypoints-based methods have been introduced, which would greatly reduce the computational cost. The two common ways in this regard are SIFT and SURF. The first keypoint-based methods \cite{huang2008detection}, \cite{amerini2010geometric} use SIFT to extract and describe keypoints. For matching, instead of using the similarity of two feature vectors, the ratio between the nearest neighbor to the second nearest neighbor is used. The challenge of keypoint-based methods is to identify smooth regions that have been tampered. It is due to the fact that sufficient keypoints are not extracted from such regions.

There are other approaches to detect copy-move forgery. A method is proposed in \cite{li2015segmentation} and \cite{lin2018region}, which divides the image into non-overlapping semantic parts and then extracts keypoints from the entire image. Matching between keypoints is accomplished. Two regions are identified as matched if they have a certain ratio of the matched keypoints. Moreover, the original and copied region should not be in the same region.

The PatchMatch algorithm introduced in 2009 \cite{barnes2009patchmatch} is a randomized algorithm that searches the entire image randomly to find the closest neighbor. Also, this algorithm has been used to detect copy-move forgery \cite{cozzolino2015efficient}. This method has less computational complexity due to utilizing a random search algorithm.

Considering the fact that employed keypoint extraction methods are not designed for copy-move detection, in \cite{zandi2016iterative}, a new method for extracting keypoints is proposed. In this way, even smooth regions are covered in an adaptive manner. 

As mentioned, methods for detecting copy-move, and splicing forgery are different. In copy-move forgery, original and copied regions are from the same image, and some image features such as color, texture, and noise are the same in both regions. As a result, methods for detecting splicing forgery cannot be used to detect copy-move forgery. The result of the copy-move detection is a binary mask in which the tampered regions are not discriminated from the original one. The similarity of the image structure in these two regions makes this discrimination too challenging. Resolving this challenge is our goal in this paper.

\section{Separation the main region of the forged region}\label{method}
\begin{figure}[htbp]
\centerline{\includegraphics[width=0.5\textwidth]{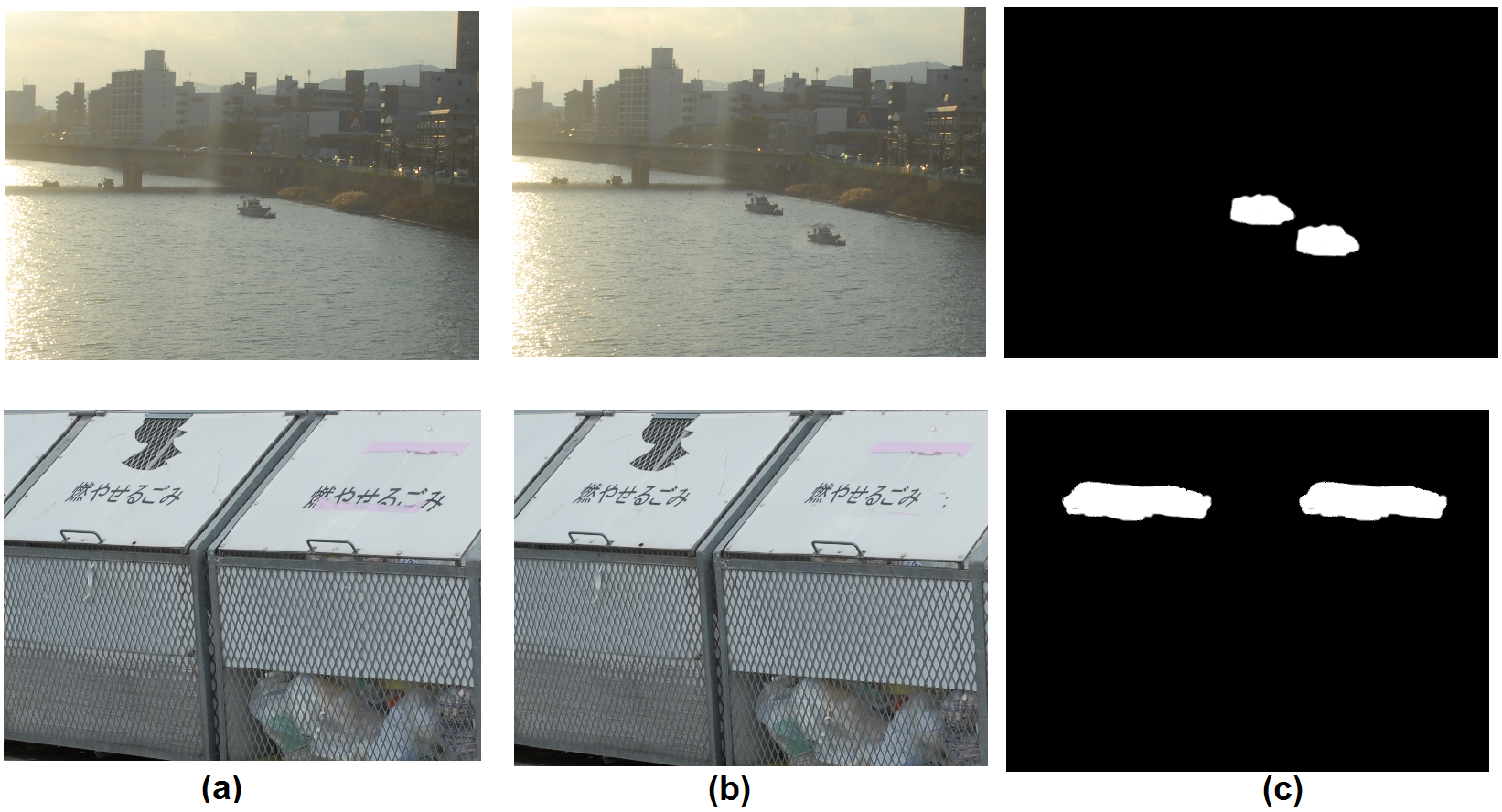}}
\caption{a) Original Image  b) Forged Image c) Binary Mask}
\label{metod1}
\end{figure}

Many techniques have been proposed to address the challenges of copy-move forgery detection. All previous methods just identify similar patches (original and copied). They are usually based on finding similarity in an image. Finally, they present a binary mask as output (like fig. \ref{metod1}). The mask does not determine which region is the original one. Identification and separation the original region among detected regions due to features similarity is a complicated task which is investigated in this paper.

In copy-move forgery, forgers usually manipulate the boundaries of the copied snippets to conceal their footprint. This intervention may lead to texture inconsistency. When these changes are performed skillfully, it is too hard to be recognized by the HVS. Therefore, it can be helpful to examine the inconsistencies in the forged regions. It should be noted that natural images are intrinsically self-similar. Therefore, it can assist a forger to conceal his footprint even without any modification. It makes it too challenging to distinguish between intact and duplicated patches.

The image texture describes the local arrangement of color and intensities. Local texture consistency might be damaged after any manipulation performed to mask the trace of forgeries. As a result, texture analysis can be exploited to discover local inconsistency.  Local binary patterns (LBP) \cite{ojala1996comparative} is kind of visual descriptor and one of texture analysis methods which generate proper features for texture classification. Since LBP extract statistical and structural features of the textures, they are considered as a powerful tool for texture analysis. They are used in many applications such as image quality assessment, face recognition, motion analysis, video and image retrieval, and so on.

In LBP, pixels’ brightness (intensity) is compared with the neighboring pixels’ brightness. Neighboring pixels can be selected in different radiuses and get a value zero or one based on differences with the central pixel. The values of neighboring pixels are converted from binary into decimal. In equation (\ref{LBP})\cite{muthevi2017leaf}, $P$ is the number of the neighborhood in radius $R$.

\begin{equation}
LBP_{P,R}(u,v) = \sum_{p=0}^{p-1}(I(g_{p}-g(u,v))2^{p}\label{LBP}
\end{equation}
In equation(\ref{LBP}) $g (u, v)$ is the intensity of the central pixel in position $(u, v)$. $g_p$, is the intensity of  its neighbors, and $I$ is calculated using (\ref{sub.LBP}):

\begin{equation}
I(x) =
\begin{cases}
1 & x\geq0 \\
0 & x<0
\end{cases}\label{sub.LBP}
\end{equation}

In order to discriminate the forged patches, LBP is applied to the grayscale image. As mentioned, the forger usually manipulates the boundaries of the forged region to eliminate the effect of tampering. Therefore, the histograms of the boundary texture of detected regions are investigated. The histogram is obtained by using (\ref{hist})\cite{muthevi2017leaf}:

\begin{equation}
hist(k) = \sum_{i}^{M}\sum_{j}^{N}, f(LBP_{P,R}(i,j),k)   , k\in[0,K] \label{hist}
\end{equation}
Where $k$ is maximum LBP pattern value. $M$, $N$  are related to interested region and the function $f$ shows the frequency each of the image texture values (\ref{sub.hist}).

\begin{equation}
f(x,y) =
\begin{cases}
1 & x=y \\
0 & \mbox{other wise }
\end{cases}\label{sub.hist}
\end{equation}

Since the forged regions are usually modified by a low pass filter in order to disappear its borders with the background, it is expected that the LBP histogram of duplicated regions will be more smooth. To check the histogram fluctuations, we employ the standard deviation as illustrated in equation (\ref{sub.s.deviation}):

\begin{equation}
s = (\frac{1}{n-1}\sum_{i=1}^{n}(hist(i)-\overline{hist})^2)^\frac{1}{2}\label{s.deviation }
\end{equation}

\begin{equation}
\overline{hist}= \frac{1}{n}\sum_{i=1}^{n} hist(i)\label{sub.s.deviation}
\end{equation}
Where $n$ is the number of elements of the histogram.
\begin{figure}[htbp]
\centerline{\includegraphics[width=0.5\textwidth]{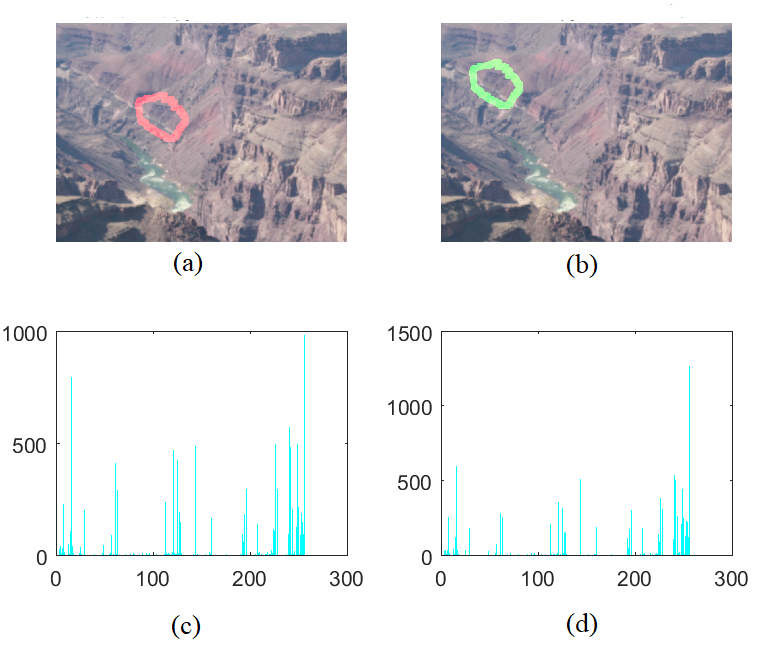}}
\caption{a) Forgery Region Boundary b) Original Region Boundary c) LBP Boundary Histogram of Detected Patch in Forgery Region d)LBP Boundary Histogram of Detected Patch in Original Region}
\label{metod4}
\end{figure}

\begin{figure*}[htbp]
\centerline{\includegraphics[width=1\textwidth]{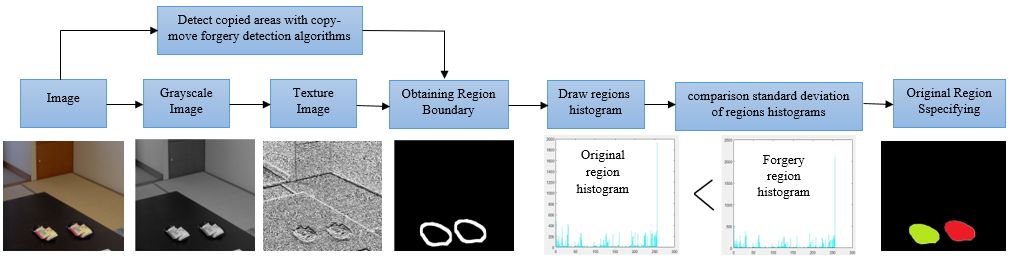}}
\caption{Proposed Method Flowchart}
\label{metod3}
\end{figure*}

\begin{table*}[htbp]
\caption{Datasets}
\begin{center}
\scalebox{0.92}{
\begin{tabular}{|c|c|c|c|c|c|c|}
\hline
\multicolumn{2}{|c|}{\textbf{Dataset}} &\textbf{Number of Image} &\textbf{Number of Original Image} &\textbf{Number of\ Forgery Image} & \textbf{Attack} & \textbf{Mask}\\
\cline{1-7}
\multicolumn{2}{|c|}{\textbf{IMD}} & 48 & 48 & 48 & Rotation, JPEG, Noise Gaussian, Scale & yes\\
\cline{1-7}
\textbf{MICC} & MICC-F220 & 220 & 110 & 110 & --- & No \\
\cline{2-7}
 & MICC-F2000 & 2000 & 1300 & 700 & --- & No\\
\cline{1-7}
\multicolumn{2}{|c|}{\textbf{SBU-CM16}} & 16 & 0 & 16 & Rotation, JPEG, Noise, Blurring & yes\\
\cline{1-7}
\multicolumn{2}{|c|}{\textbf{GRIP}} & 80 & 80 & 80 & --- & yes\\
\cline{1-7}
\hline
\end{tabular}
}
\label{tab1}
\end{center}
\end{table*} 

Thus, by calculating the standard deviation of the LBP histogram, it is possible to detect the copied patches. In other words, the standard deviation of the LBP histogram is expected to be less than its counterpart in original snippet.
The LBP histogram of a boundary of a forged and an original one is compared in fig.\ref{metod4}. The flowchart of the proposed method is illustrated in fig.\ref{metod3}. 

\section{Experimental Results}\label{result}
In this section, the evaluation results of the proposed method are presented. In subsection \ref{result.data}, the available benchmark datasets are explored.

\subsection{Dataset}\label{result.data}
There are several benchmark datasets to evaluate copy-move forgery detection. Some of them are not suitable for evaluating the proposed method since they are provided automatically. Consequently, they are easily recognizable by the HVS. In this research, we use the GRIP dataset  \cite{cozzolino2015efficient}, which is prepared more skillfully.

The MICC datasets are constructed by Amerini et al.\cite{amerini2011sift}. The first one, MICC-F220 includes 220 images which 110 of them are originals and others are tampered images. The second one, MICC-F2000 includes 2000 images which 1300 of them are originals and others are tampered images. In this dataset, the ground truth masks are not provided. 

The Image Manipulation Dataset (IMD) is provided by Christine et al. \cite{christlein2012evaluation}. This dataset contains 48 images. It generally contains high resolution images. Rotation, scale, compression and Gaussian noise in varying degrees are also applied on duplicated regions.

The SBU-CM16 dataset is prepared by Zandi et al. \cite{zandi2016iterative}. This collection of images is made based on 16 images. These images are subjected to four kinds of attacks such as Gaussian noise, compression, blur and rotation with varying degrees. The images size is approximately $800\times580$. Employing overly smooth images makes it a challenging benchmark dataset. However, the forged regions are not prepared skillfully.

\begin{figure*}[htbp]
\centerline{\includegraphics[width=1\textwidth]{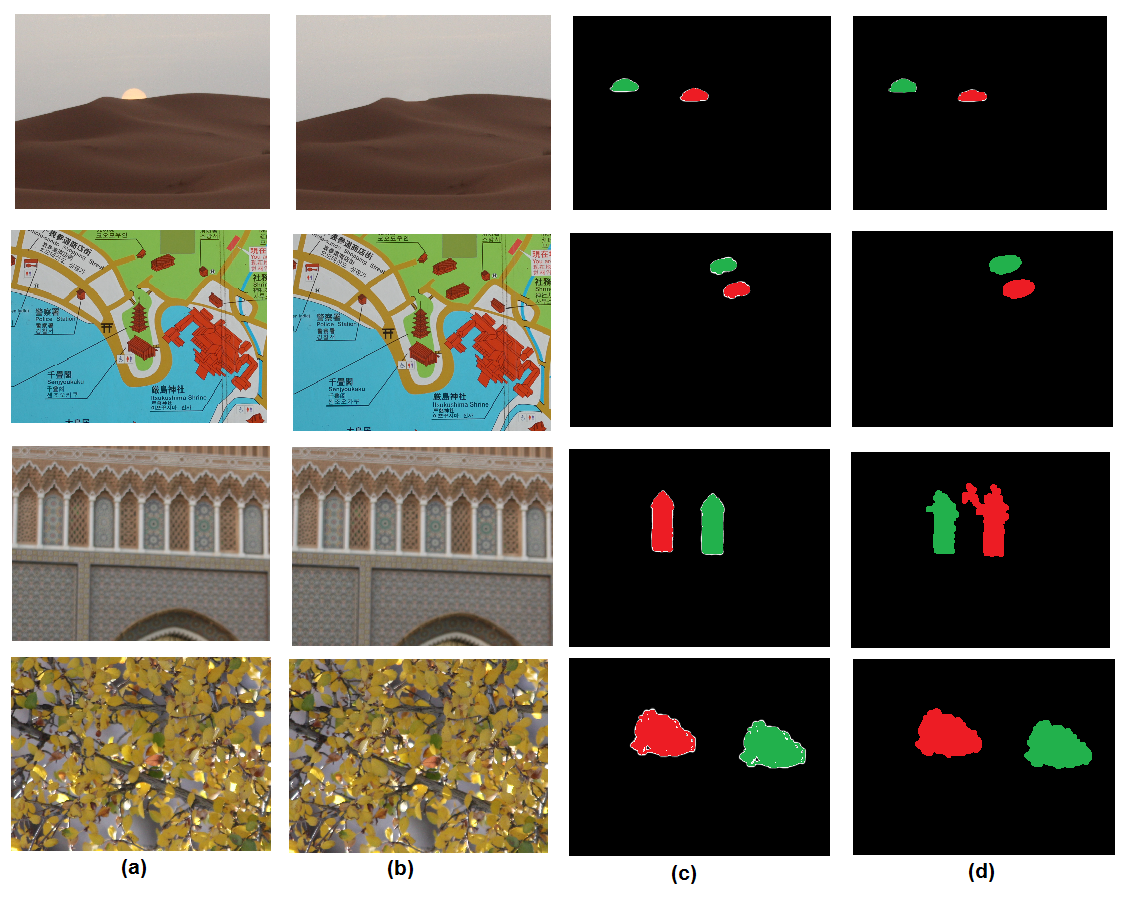}}
\caption{Proposed Method Qualitative Evaluation:
a) Original Image b) Forged Image c) Binary Mask d) Final Result in Detected Mask }
\label{metod5}
\end{figure*}

The GRIP dataset is presented by Cozzolino et al. \cite{cozzolino2015efficient}. This dataset contains 80 images at 1024 × 768, and the copied region has different shapes and sizes. Copy area size of this dataset is approximately 4,000 (less than 1 percent of the whole image) to 5000 pixels.

In this paper, we find out the GRIP dataset is appropriate for evaluating the proposed method because we are looking for forgery images that are not recognizable by the HVS. The other datasets which introduced are usually provided automatically and are easily recognizable by the HVS.

\subsection{Performance evaluation}\label{result.evaluation}
LBP uses different radiuses to select neighborhoods. Since the forger usually conceals his trace by using filters which are proportional to image resolution and forged area; as a result, we consider a range corresponding to the above mentioned points for analyzing the image texture. This range of radius is adopted from \cite{wang2004image}, which is used to investigate the image quality and provides an appropriate scale based on the resolution of the image. It is needless to say that the forgers should be considerd the characteristic of HVS in order to forge an image in a believable manner. Because the size of the images in the GRIP dataset approximately are the same, the three radiuses, depicted in Table \ref{tab2} are used for LBP. In general, if inconsistency of texture is detected in two radiuses of three, it can be concluded that the region is forged. In our implementation, the number of neighbors is considered as 8 (P = 8) in LBP.

We apply the proposed method twice on the GRIP dataset images. Once we use the exact masks(ground-truth) of the regions to find out boundaries of the original and forged area for separating them. Since we are looking for a completely automatic and practical way to localize the forged patches, 
the resultant mask of the PatchMatch which is one of the best methods is exploited as well.

As it is shown in table \ref{tab2}, this method can separate the original from the forged area with 67.5 percent accuracy. Since our assumption is that the forger has manipulated the boundaries of the forged area, this method works well where it needs to be manipulated to hide tampering trace. Nonetheless, it is possible to forge an image without any manipulation due to inherent self-similarity of natural images. Moreover, high texture regions can be easily subjected to copy-mover forgery without any try to conceal the trace. In this case, the proposed method cannot detect the forged patches correctly. An example of such images are depicted in fig.\ref{metod5}. 
\begin{table}[htbp]
\caption{The experimental results on the GRIP}
\begin{center}
\begin{tabular}{|c|c|c|c|c|}
\hline
&\textbf{R=2} & \textbf{R=3} & \textbf{R=4} & \textbf{Final Result} \T\\
\cline{1-5}
\textbf{Grand-truth} & 65\% & 66\% & 67.5\% & 67.5\% \T\\
\cline{1-5}
\textbf{Detected Mask} & 66\% & 70\% & 70\% & 70\%\T \\
\hline
\end{tabular}
\label{tab2}
\end{center}
\end{table}

\section{Conclusion}\label{Conclusion}
Copy-move forgery detection methods are mostly based on finding similar regions. They provide a binary mask as their output in which each pixel is identified as either background or copy-move pixels. In this paper, a method for discriminating the duplicated region from the original one is presented. Our method employs texture information of the border regions of detected copy-move regions. Since the original and forged region is parts of the same image, detecting the duplicated snippet is a challenging task. The proposed method has been validated using GRIP dataset. the presented method can detect the forged regions with accuracy of 67.5\%.
\bibliographystyle{ieeetr}
\bibliography{refrence}

\end{document}